\documentclass[11pt,a4paper]{article}

\usepackage{graphicx}
\usepackage{booktabs}

\usepackage{fancyhdr}
\fancypagestyle{firstPageStyle}
{
   \fancyhf{}
        \chead{}
        \lfoot{ \fontsize{10pt}{1em}\selectfont pp. 37-46, 2020. CS \& IT - CSCP 2020}
        \rfoot{ \fontsize{10pt}{1em}\selectfont DOI: 10.5121/csit.2020.101803}
	\cfoot{}
	\lhead{ \fontsize{10pt}{1em}\selectfont David C. Wyld et al. (Eds): CSEA, DMDBS, NSEC, NETWORKS, Fuzzy, NATL, SIGEM - 2020}
	\rhead{}
}

\usepackage{titlesec}

\titlespacing*{\section}{0pt}{0pt}{0pt}
\titlespacing*{\subsection}{0pt}{4pt}{4pt} 
\titleformat{\section}{\fontsize{14pt}{1em}\selectfont\bfseries\scshape}{\thesection.}{1em}{}

\usepackage[pagebackref=true]{hyperref}
\hypersetup{
    pdftoolbar=true,        
    pdftitle={Minimum Viable Model Estimates for Machine Learning Projects},    
    pdfauthor={John Hawkins},     
    pdfsubject={Minimum Viable Model Estimates for Machine Learning Projects},   
    pdfkeywords={Machine Learning} {ROI Estimation} {Machine Learning Metrics} {Cost Sensitive Learning}, 
    colorlinks=false}

\begin{document}


\thispagestyle{firstPageStyle}

{ \vspace*{-0.05cm} \fontsize{21pt}{2em}\selectfont
  \centering Minimum Viable Model Estimates for Machine Learning Projects \par \vspace*{24pt minus \parskip} }
%
%
{ \centering  \fontsize{13pt}{1em}\selectfont John Hawkins \par\vspace*{12pt minus \parskip}}
{ \centering  \fontsize{13pt}{1em}\selectfont Transitional AI Research Group, Sydney, Australia\par \vspace*{18pt minus \parskip}}
{ \centering  \fontsize{10pt}{1em}\selectfont john@getting-data-science-done.com \par\vspace*{18pt minus \parskip} }

{ \fontsize{14pt}{1em}\selectfont \textbf{\textit{Abstract}} } \par \vspace*{6pt minus \parskip}

{  \fontsize{10pt}{1em}\selectfont
\textit{
Prioritization of machine learning projects requires estimates of both the potential ROI
of the business case and the technical difficulty of building a model with the required
characteristics.
In this work we present a technique for estimating the minimum required
performance characteristics of a predictive model given a set of information about
how it will be used. This technique will result in robust, objective comparisons between potential projects.
The resulting estimates will allow data scientists and managers
to evaluate whether a proposed machine learning project is likely to succeed
before any modelling needs to be done.
The technique has been implemented
into the open source application MinViME (Minimum Viable Model Estimator)
which can be installed via the PyPI python package management system, or downloaded directly from
the GitHub repository.
Available at \href{https://github.com/john-hawkins/MinViME}{https://github.com/john-hawkins/MinViME}
} \par \vspace*{6pt minus \parskip} }
 
{ \fontsize{14pt}{1em}\selectfont \textbf{\textit{Keywords}} } \par \vspace*{6pt minus \parskip}
{ \fontsize{10pt}{1em}\selectfont \textsl{Artificial Intelligence, ROI Estimation, Machine Learning Metrics, Cost Sensitive Learning} }\par \vspace*{18pt minus \parskip}
\section{Introduction}
%
\titlespacing*{\section}{0pt}{8pt}{6pt}
In an ideal world we would priortise all our machine learning projects according to their expected payoff.
This would, in turn, require a reasonable estimate of both the probability of success and the return given
success. Creating such an estimate is difficult and instead we tend to be guided by heuristics,
implicit and explicit, about the size of the problem and its difficulty.

Difficulty estimation is often limited to a discussion about what
it would take to feasibly productionise a model.
Consideration of the difficulty of building the model, to the extent that it is done at all, is usually
an informal process relying on the experience of data scientist working on the project.
In this work we present a structured approach to estimating the difficulty of a machine learning
task by estimating the minimum required performance of a model that would meet the business objectives.

The most common approach to which the details of the business case are considered in the building of a machine
learning solution is through development of domain specific methods. For example, work on designing
loss functions for specific applications (See \cite{Johnson19}, \cite{Hennig2007}). Alternatively, there is a
practice of using cost sensitive learning \cite{Domingos1999,Margineantu2000,Elkan2001,Tian+Zhang2019}
to permit general techniques learn a solution that is tuned to the
specifics of business problem (for example \cite{Fatlawi2017}). In general, you can achieve
the same outcome by learning a well calibrated probability estimator and adjusting the decision threshold
on the basis of the cost matrix \cite{Nikolaou2016}.
The cost matrix approach to analysing a binary classifier can be also be used to estimate its expected ROI \cite{Ylijoki2018}.

These approaches allow you
to improve and customise a model for a given application, but they do not help you decide if the project is
economically viable in the first place. There have been extensive efforts to characterise project
complexity in terms of data set requirements for given performance criteria \cite{Raudys1991}, however the
task of determining these criteria from business requirements has not, to our knowledge, been addressed.
Fortunately, as we will demonstrate, some of the machinery for the cost matrix
analysis can be used in reverse to estimate the baseline performance a model would require to reach
minimal ROI expectations.

We use the framework of a cost matrix to develop a method for estimating the minimum viable binary classification
model that satisifies a quantitative definition of the business requirements. We demonstrate that a lower bound
on \textit{precision} can be calculated \textit{a priori}, but that most other metrics require knowledge of either the expected number
of false positives or false negatives. We devise an additional \textit{simplicity} metric which allows for direct
comparison between machine learning projects purely on the basis of the business criteria. Finally, we demonstrate
that lower bounds on the AUC and other metrics can be estimated through numerical simulation of the ROC plot.

We use the minvime ROC simulator to conduct a range of experiments to explore the manner in which the minimum viable
model changes with the dimensions of the problem landscape.
The results of these simulations are presented in a manner to help develop
intuitions about the feasibility of machine learning projects.

\section{Global Criteria}\label{sec:global}

In order to estimate the required performance characteristics of any machine learning model there are several
global criteria that need to be defined.
Firstly, we need to determine the overall ROI required to justify the work to
build and deploy the model. This can take the form of a project cost plus margin for expected return.
Given an overall estimate we then need to amortise it to the level of analysis we are working at,
for example, expected yearly return.

In addition, we require the expected number of events that will be processed by the machine learning system for
the level of analysis mentioned above. For example, this may be the number of customers that we need to analyse
for likelihoood of churn every year. Included in
this we require to know the \textit{base rate} of the event we are predicting. In other words how many people tend to churn
in a year.

Given the above criteria, we next need to evaluate how a model would need to perform in order to process the events
at a sufficient level of compentence. To do this analysis
we need to look at the performance of the model as it operates on the events we are predicting. We start this
process by looking at the impact a binary classifcation model using a cost matrix analysis.

\section{Cost Matrix Analysis}

The idea of a cost matrix in machine learning emerged from the literature on training models on imbalanced datasets. Specifically
there are a set of papers in which the learning algorithm itself is designed such that the prediction returned is the one that
minimises the expected cost \cite{Elkan2001,Margineantu2000}.
Typically this means not necessarily returning the class that is most probable, but the one with the
lowest expected cost. In essence, the model is learning a threshold in the probability space for making class
determinations. The same idea can be applied after training an arbitrary classfier that returns a real-value result
in order to determine a threshold that minimises the error when the model is applied.

It has been observed that the later approach is not dissimilar from a game theoretical analysis \cite{Sanchez17}.
In game theory the cost matrix would delineate the outcomes for all players of a game in which each player's fortunes
are affected by their own decision and that of the other player. Each axis of the matrix describes the decisions of a player
and by locating the cell at the nexus of all decisions we can determine the outcome of the game.

In binary classification we replace the players of the game with two sources of information: the prediction of the model and
the ground truth offered up by reality. For binary classification problems this is a 2x2 matrix in which each of the cells have
a direct corollary in the confusion matrix. In the confusion matrix we are presented with the number of events that fall into
each of the four potential outcomes. In a cost matrix we capture the economic (or other) impact of each specific type of event
occuring. From here on in we will follow Elkan \cite{Elkan2001} and allow the cells of the matrix to be either depicted as
costs (negative quantities) or benefits (positive quantities).

In table \ref{costmatrix} we demonstrate the structure of a cost matrix in which we also depict the standard polarity of the costs and
benefits in most applications.

\begin{table}[htbp]
\centering
 \caption{Cost/Benefit Matrix for a Binary Classification System.}
 \label{costmatrix}
  \begin{tabular}{|l||l|l|}
   \toprule
                          &Actual Negative              &Actual Positive       \\
   \midrule
   Predict Negative       &True Negative (TN)           &False Negative (FN)           \\
   Predict Positive       &False Positive (FP) Cost     &True Positive (TP) Benefit    \\
   \bottomrule
  \end{tabular}
\end{table}

If we assume, without loss of generality, that the positive class corresponds to the event against which we
plan to take action, then as observed by Elkan \cite{Elkan2001} the entries of the cost matrix for the predicted
negative row should generally be identical. The impact of taking no action
is usually the same. It is also usually zero because the model is typically being applied
to a \textit{status quo} in which no action is taken.
In most cases you will only need to estimate the cost/benefits of your true positives
and false positives. This is because the assumed ground state (the status quo) is the
one in which the actual costs of false negatives are already being incurred. Similarly
the true negatives are the cases in which you have taken no action on an event that
does not even represent a loss of opportunity.

The overall outcome of using a model can be calculated by multiplying the cells of the confusion matrix by the corresponding
values in the cost matrix and summing the results. This tells us what the overall impact would have been had we used the model
on the sample of data used to determine the confusion matrix. The advantage of using a cost/benefit structure in the matrix is
that we can then read off whether the result of using the model was net-positive or net-negative depending on the polarity of
the total impact. This approach can then be used to optimise a decision threshold
by varying the threshold and recalculating the expected impact.

Before continuing with the analysis of the cost matrix we make some observations about the process of determining the content
of the cost matrix.

\subsection{Deterministic Outcomes}

In some situations we can say in advance what the impact of a specific conjunction of prediction
and action will be. We might know, for example, that giving a loan to a customer who will default will result
in a loss of money. If we take the action of rejecting that loan then the money is no longer lost.

This determinism in outcomes is typically true of situations where once we have taken our action there is no dependence
on additional decisions by other people to impact whether our action is effective.
This is not always true. There are certain circumstances in which knowing what will happen does
not guarantee a given outcome. In these cases there is stochastic relationship between the prediction and the outcome.

\subsection{Stochastic Outcomes}

In the stochastic situation we understand that even in the event of possessing a perfect oracle for the future events,
we would still be left with an element of randomness in the outcomes of our actions. This is typically true where the action we take
will involve third parties or processes beyond our control to achieve their result.

The canonical example of a predictive process with a stochastic outcome is a customer churn model.
A churn model is used to intervene and attempt to prevent a customer from churning.
However, our ability to influence a customer to not churn is not guaranteed by accurately predicting it.

In these situations we need to make additional estimates in order to get the contents of the cost matrix. For example,
if we expect that our interventions will succeed one in every five attempts, and the value of a successful intervention
is \$1,000, then the benefit of a true positive is \$200.
Defining a cost/benefit matrix for a situation with stochastic outcomes will
require additional assumptions.

\subsection{Metrics from the Matrix}

If we have an existing model then we can use the cost matrix to estimate the economic
impact by combining it with a confusion matrix. How then can we estimate the required properties of
a model such that it renders a project viable?

Once the model is deployed it will make predictions for the $N$ cases that occur in the period of analysis.
We can demarcate the content of the unknown confusion matrix with
the symbol $tp$ true positives and $fp$ for false positives.
The benefit of each true positive is captured
by the value $\mathcal{B}$ from our cost matrix, similarly the cost of each false positive is
demarcted $\mathcal{C}$.
In order to satisfy that the model meets our ROI minimum $\mathcal{M}$ then the following
will need to hold:

\begin{equation}
\label{eq:tpfp}
tp \cdot \mathcal{B} - fp \cdot \mathcal{C} >= \mathcal{M}
\end{equation}

From Equation \ref{eq:tpfp} we can derive explicit defintions for both the number of true positive and false
positive predictions that would satisfy our minimal criteria.

\begin{equation}
tp >= \frac{fp \cdot \mathcal{C} + \mathcal{M}}{\mathcal{B}}
\end{equation}

\begin{equation}
fp <= \frac{tp \cdot \mathcal{B} - \mathcal{M}}{\mathcal{C}}
\end{equation}

Note, that these expressions involve a reciprocal relationship between the true positives and false positives.
Moderated by the ratio between the costs and benefits of taking action in relation to the overall required return.
We would like to be able to estimate the True Positive Rate (TPR) or \textit{recall},
and the False Positive Rate (FPR) or \textit{fall-out}.
To define these we need to introduce one of the other global terms required to estimate
model requirements: the base rate $r$ at which the event we are predicting occurs.
Which allows us to define TPR and FPR as follows:

\begin{equation}
TPR = \frac{ tp }{ N \cdot r }
\end{equation}
\begin{equation}
FPR = \frac{ fp }{ N \cdot (1-r) }
\end{equation}

One of the most important metrics for evaluating a binary classification model is precision (or Positive Predictive Value). This determines
the proportion of all predicted positive events that were correct. We can now generate an expression for what this would
be at the exact point of meeting of ROI requirements.

\begin{equation}
\label{eq:plb}
precision = \frac{1}{1 + \frac{\mathcal{B}}{\mathcal{C}} - \frac{\mathcal{M}}{tp \cdot \mathcal{C}}}
\end{equation}

Unfortunately, the presence of the number of true positives in this expression resists additional simplification.
However, if we focus our attention on the \textit{break even} point we can eliminate this component and derive the
following lower bound on the precision of the model.

\begin{equation}
precision > \frac{1}{1 + \frac{\mathcal{B}}{\mathcal{C}} }
\end{equation}

\begin{figure}[h!]
\includegraphics[scale=0.5]{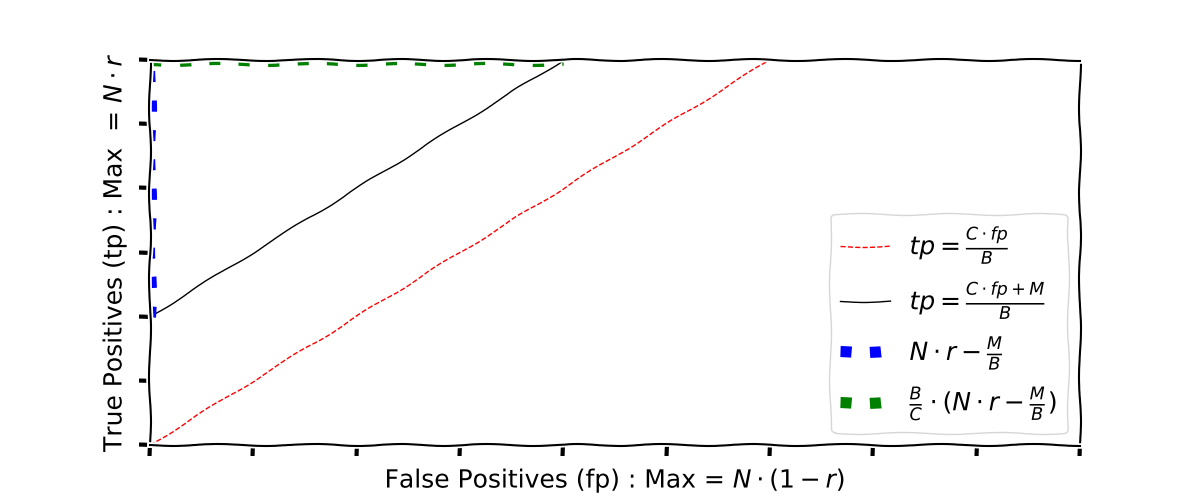}
\caption{Schematic representation of potential model outputs depicting the boundary between viable and inviable models.}
\label{fig:simplicity}
\end{figure}

This lower bound corresponds to the line on which the benefits of the true positives are counteracted by the cost of the false positives.
Equation \ref{eq:plb} corresponds to the line that lies $\frac{\mathcal{M}}{\mathcal{B}}$ units above this lower bound.
We depict both of these lines schematically in Figure \ref{fig:simplicity}.

This geometric representation suggests an additional metric for the feasibility of any given project.
Note, that the triangle in the upper left corner represents all possible model outputs that would satisfy
the stated criteria. The dimensions of this triangle are shown in the legend by the lines of blue and green squares.
We can calculate the area of this triangle and express it as a proportion of the total area of the
rectangle. 
This will result in a metric which is the proportion of all potential
model outputs that are economically viable.
We can define an expression for this ratio as shown in Equation \ref{eq:simplicity}.

\begin{equation}
simplicity = \frac{\mathcal{B}}{2 \cdot \mathcal{C} \cdot N^2 \cdot r(1-r)} \cdot  \left(N \cdot r - \frac{\mathcal{M}}{\mathcal{B}} \right)
\label{eq:simplicity}
\end{equation}

This simplicity measure allows us to rank the difficulty of problems based purely on their business criteria. 
As this metric is to be
calculated in a discrete space, and it must deal with edge cases in which the triangle is truncated, we refer 
the reader to the source code for its complete calculation.

Our efforts to estimate additional requirements in terms of standard machine learning metrics are hindered by 
the requirement to know
either the number of true positives or false positives in the minimal viable model.
To move beyond this limitation we will employ numerical techniques to estimate the bounds on the model performance.

\subsection{ROC Simulation}

\begin{figure}[h!]
\includegraphics[scale=0.5]{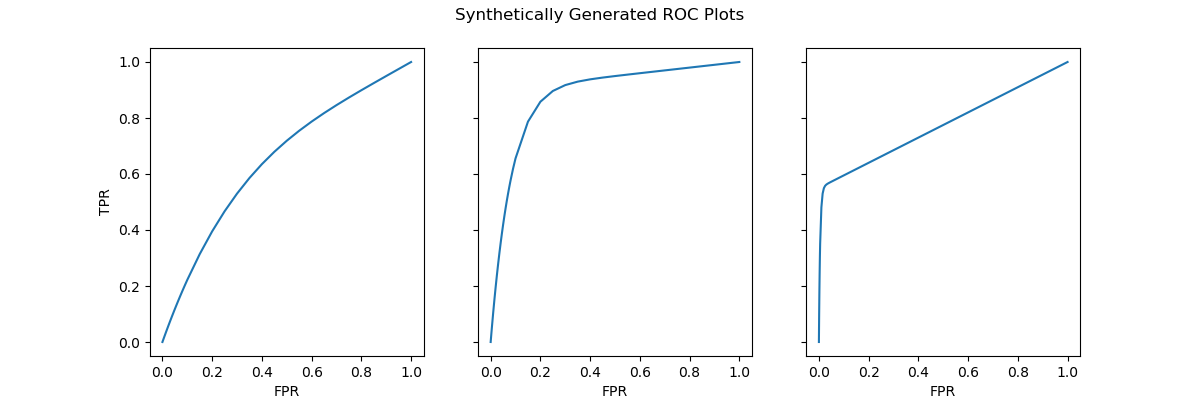}
\caption{Examples of ROC plots generated through simulation.}
\label{fig:ROC}
\end{figure}

The ROC plot provides us with a method of examinining the performance of a binary classification model
in a manner that does not depend on a specific threshold. As shown in the example plot \ref{fig:ROC}
it depicts a range of TPR and FPR values across all the different potential thresholds of the model.
Additionally, the ROC plot permits the definition of a non-threshold specific metric to evaluate the overall
discriminative power of the model (regardless of chosen threshold), the Area Under the Curve (AUC) \cite{Bradley97}.

We can generate synthetic ROC plots by exploiting their key properties.

\begin{enumerate}
        \item They are necessarily monotonically increasing
        \item They will necessarily pass through the points (0,0) and (1,1)
        \item They remain above or on the diagonal between these points (unless the model is inversely calibrated)
\end{enumerate}

We can simulate such curves by designing a parameterised function that satisfies these
criteria. We have designed the function shown in equation \ref{auc:sim} for this purpose.
It is composed of two parts, both of which independantly satisfy the criteria.

\begin{equation}
\label{auc:sim}
y = \alpha \cdot (-(x-1)^{2\beta}+1) + (1-\alpha) \cdot x
\end{equation}

In this function $\alpha$ is a weighting between zero and one that determines the
mix between the two component functions. The first component function is an
inverted even-powered polynomial function of x, offset so that its origin
lies at coordinates {(1,1)}.
The second of function is just $x$ itself, which is the diagonal line.
Both functions necessarily pass through
the points $(0,0)$ and $(1,1)$, and their mixture determines the shape of the
curve between these points.

Figure ~\ref{fig:ROC} shows some examples of simulated ROC plots with different values of
the parameters $\alpha$ and $\beta$. Note, it is critical to restrict the space of these parameters such
that the search space does not result in unrealistic ROC plots. The third plot in Figure ~\ref{fig:ROC}
demonstrates that at extreme values the curve starts to represent a spline of two linear functions.

By running simulations across a variety of values of $\alpha$ and $\beta$
we can generate multiple synthetic ROC plots. Each plot provides a set of hypothetical
values for the true positive rate and false positive rate across a range of thresholds.
If we combine these with the pre-defined characteristics of the problem we are solving
(the number of cases, baseline rate, costs and benefits). Then we can identify which
of these synthetic ROC plots would represent acceptable models. We can then rank
them such that we search for the ROC curve with the minimum AUC that satisfies our
ROI constraints.

Given a minimum viable AUC, we can then calculate estimates for the other metrics discussed
above by taking the TPR and FPR values at the minimum viable threshold.

We have built a Python tool for running such simulations and released it as a Python package
to the PyPI library (Available at $https://github.com/john-hawkins/MinViME$).

\section{Simulations}

The MinViME tool allows us to explore the viable model landscape across a broad range of business
criteria in order to understand what drives feasibility. In order to do so we need to represent
the space of business problems in an informative and comprehensible way. We do so using the following
three ratios.

\begin{figure}[!ht]
\includegraphics[scale=0.5]{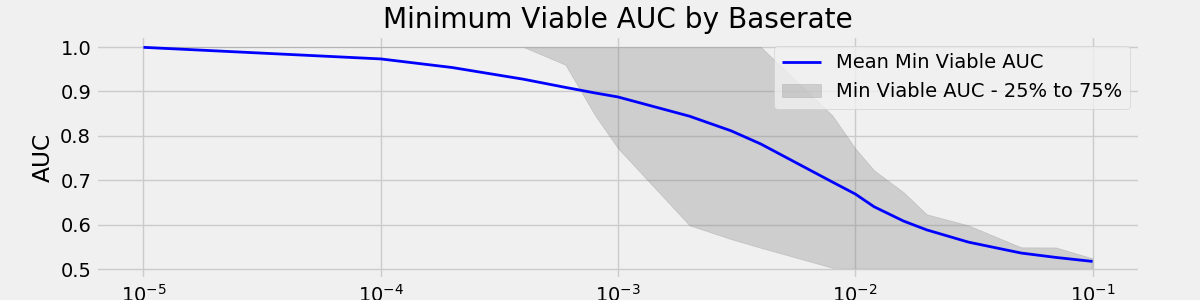}
\includegraphics[scale=0.5]{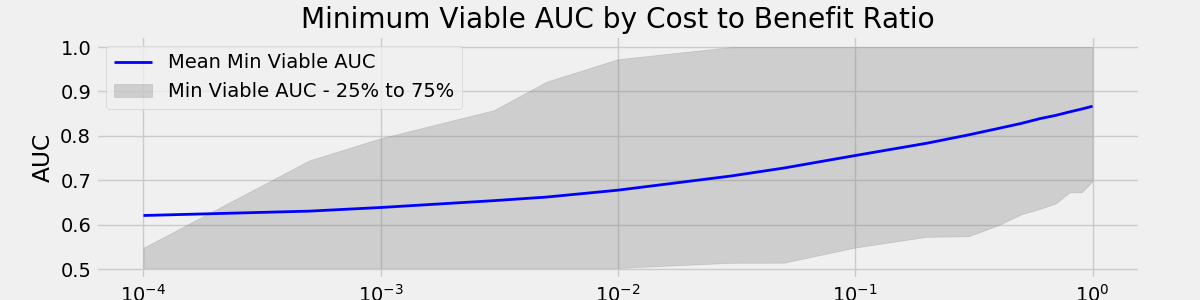}
\includegraphics[scale=0.5]{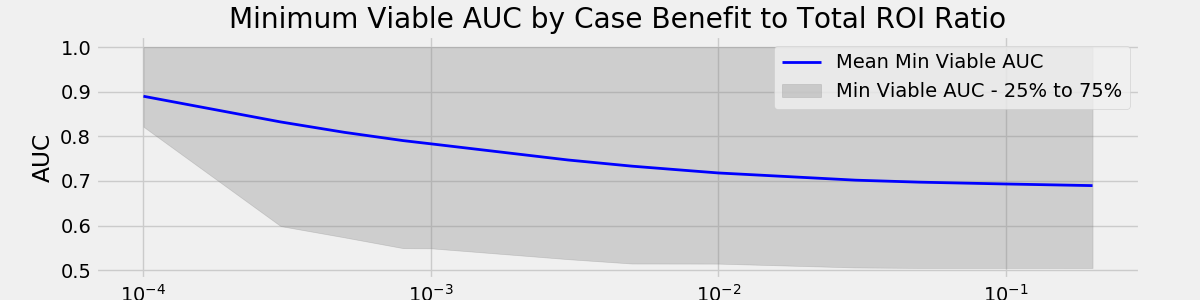}
\caption{Relationship between the AUC of the minimum viable model and the criteria that determine
the business problem. Note: We use a log scale on the X axis to demonstrate how
the difficulty varies with each order of magnitude change in the nature of the problem.
}
\label{fig:threeplots}
\end{figure}

\begin{enumerate}
        \item \textbf{Benefit to Total ROI.} In other words how significant to the total required ROI is
each additional true positive. We explore a range of values between zero and one.

        \item \textbf{Cost to Benefit.} In other words how many false positives can we tolerate for every
true positive case we identify. In principle this could take values greater than one, however this
is almost never seen in practice so we explore a range of values between zero and one

        \item \textbf{Base Rate.} The rate at which the event being predicted occurs in the population.

\end{enumerate}

In addition, we use a static value of one million for the number of cases processed per unit of time
This value is realistic for many problems and our experimentation with values
an order of magnitude either side demonstrates little variation on the results presented below.

\subsection{Results}

In Figure \ref{fig:threeplots} we see the mean minimum required AUC across each of the three dimensions.
The gray bounds indicate the lower and upper quartiles. We see that problems get easier as either the base
rate or the benefit to total ROI increases. The opposite occurs for the cost to benefit ratio as we would
expect. The base rate plot demonstrates the most variability across the spectrum, with base rates between
$10^{-3}$ and $10^{-2}$ demonstrating the largest variability. Above this range problems generally have less
strenuous requirements and below it most problems require demanding model performance.

Elkan \cite{Elkan2001} observes that if we scale the entries in a cost matrix by a positive constant or if
we add a constant to all entries, it will not affect the optimal decision. However, this
is not true for determining the performance characterists of the minimum viable model.
The interaction between the number of cases, the required ROI and the values in the cost
matrix will mean that these changes can affect the characteristics of the minimum viable model.

When the benefit to total ROI ratio is very small we see that the model requirements are predominantly
for an AUC above $0.8$. To further understand the relationship between the base rate and the cost to benefit ratio
we plot the surface of required AUC for these values when the benefit to total ROI is $10^{-4}$.

\begin{figure}[h!]
\includegraphics[scale=0.5]{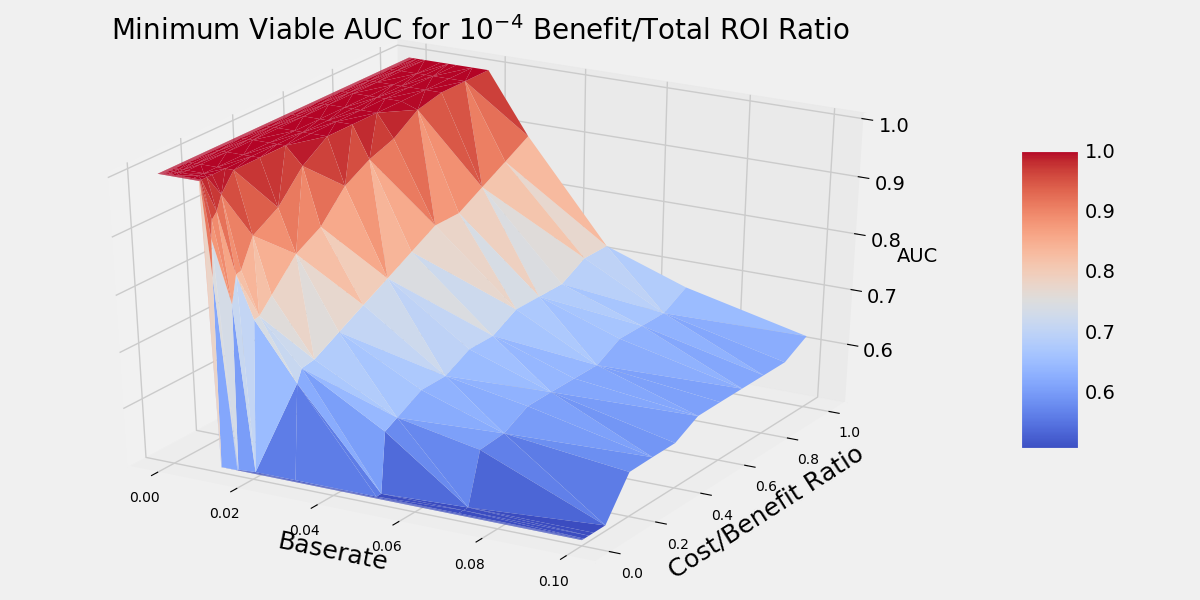}
\caption{Surface of the AUC landscape for business problems in which each TP prediction returns
$10^{-4}$ of the total required ROI}
\label{fig:CBRatio}
\end{figure}

In Plot \ref{fig:CBRatio} we can see clearly that very low base rates can generate intractable problems that
are likely to extremely difficult to solve. As the base rate increases we see a general weakening of the
effect of the cost/benefit ratio. However, at very low cost/benefit ratios the impact of the base rate
is almost entirely mitigated.

\section{Conclusion}

We have discussed the problem of estimating the minimum viable performance characteristics
of a machine learning model. The goal is to be able to do this before projects are
undertaken so that we can make informed decisions about where our resources are
most effectively deployed.

We investigated whether analytical lower bounds of standard machine learning performance metrics could be calculated
from business criteria alone. We derived this for the model \textit{precision}, and derived a novel \textit{simplicity}
metric which permits \textit{a priori} comparisons of project complexity for binary classification systems.

We then demonstrated a numerical approach to estimating the minimum viable model performance through simulation
of ROC plots. This in turn allows lower bound estimation of all other standard binary classification metrics.
Using this method we explored the space of minimum viable models for a range of business problem characteristics.
What we observe are some critical non-linear
relationships between the base rate and the cost/benefit ratio that will determine whether a project
is feasible or likely to be intractable.

\bibliography{refs}{}
\bibliographystyle{IEEEtran}

\hfill \break
\hfill \break
\hfill \break
 \fontsize{10pt}{1em}\selectfont © 2020 By AIRCC Publishing Corporation. This article is published under the Creative Commons Attribution(CC BY) license.

\end{document}